# Predicting Student Dropout Risk With A Dual-Modal Abrupt Behavioral Changes Approach

Jiabei Cheng
*Department of Computing*
*The Hong Kong Polytechnic University*
Hong Kong, China
jiabcheng@polyu.edu.hk

Zhen-Qun Yang*
*Department of Computing*
*The Hong Kong Polytechnic University*
Hong Kong, China
zq-cs.yang@polyu.edu.hk

Jiannong Cao
*Department of Computing*
*The Hong Kong Polytechnic University*
Hong Kong, China
jiannong.cao@polyu.edu.hk

Yu Yang
*Centre for Learning, Teaching, and Technology*
*The Education University of Hong Kong*
Hong Kong, China
yangyy@eduhk.hk

Xinzhe Zheng
*Department of Computing*
*The Hong Kong Polytechnic University*
Hong Kong, China
johnnyzheng0636@gmail.com

*Abstract*—Timely prediction of students at high risk of dropout is critical for early intervention and improving educational outcomes. However, in offline educational settings, poor data quality, limited scale, and high heterogeneity often hinder the application of advanced machine learning models. Furthermore, while educational theories provide valuable insights into dropout phenomena, the lack of quantifiable metrics for key indicators limits their use in data-driven modeling. Through data analysis and a review of educational literature, we identified abrupt changes in student behavior as key early signals of dropout risk. To address this, we propose the Dual-Modal Multiscale Sliding Window (DMSW) Model, which integrates academic performance and behavioral data to dynamically capture behavior patterns using minimal data. The DMSW model improves prediction accuracy by 15% compared to traditional methods, enabling educators to identify high-risk students earlier, provide timely support, and foster a more inclusive learning environment. Our analysis highlights key behavior patterns, offering practical insights for preventive strategies and tailored support. These findings bridge the gap between theory and practice in dropout prediction, giving educators an innovative tool to enhance student retention and outcomes.

## I. INTRODUCTION

PREDICTING student behavior has always been a significant topic in educational research, with dropout prediction being particularly crucial. Effective dropout prediction not only aids in school management but also provides timely support to students and their families, preventing potential problems. The importance of this topic lies not only in saving students who might drop out but also in improving overall educational quality and equity. Therefore, there is an urgent need to develop precise methods for predicting dropout behavior.

Methods for predicting student dropout behavior have evolved through several stages, each with different approaches and focuses. Early methods relied mainly on simple statistical analysis, typically focusing on basic demographic and academic performance data. For example, Tinto (1975) analyzed the impact of student social and academic integration on dropout rates, initially exploring the reasons for student dropout [1]. The advantage of this method is its simplicity and ease of operation, but it fails to capture complex behavior patterns and dynamic changes. Bean (1980) further developed Tinto's model, proposing an economic theory-based student dropout model, suggesting that student dropout results from multiple interacting factors [2]. Despite providing a more indepth explanation of dropout reasons, Bean's model still relied on static data and could not dynamically track changes in student behavior.

With the advancement of computer technology, researchers began to adopt traditional machine learning methods, using more complex models to predict student dropout behavior. Baker and Yacef (2009) proposed using student behavior data for prediction, applying decision trees and random forest algorithms to improve prediction accuracy [3]. This approach offers the advantage of utilizing more feature data, but these models still rely on extensive feature engineering work. Dekker et al. (2009) used the Support Vector Machine (SVM) [4] to predict student dropout behavior, further enhancing model performance [5]. However, these traditional machine learning methods often require a large amount of manual feature engineering and struggle to capture the dynamic changes in student behavior.

The introduction of deep learning technologies marked a significant leap in this field. Deng et al. (2017) proposed a method based on Convolutional Neural Networks (CNNs) [6], which significantly improved prediction performance by automatically extracting features [7]. Deep learning models can handle larger datasets and capture complex patterns within the data. Goodfellow et al. (2016) further applied Recurrent Neural Networks (RNNs) [8] to student behavior prediction, enabling the handling of sequential data and capturing dynamic changes in student behavior [9]. With the development of computational capabilities and the emphasis on deep learning technologies, researchers focus on multimodal data fusion methods [10], [11]. In the education

This work is supported by Hong Kong Jockey Club Charities Trust (Project S/N Ref.: 2021-0369), and the Research Institute for Artificial Intelligence of Things, The Hong Kong Polytechnic University (Corresponding author: Zhen-Qun Yang).

field, researchers are starting to focus on the application of online education data, providing a more holistic student profile by integrating academic performance, behavioral data, and psychological traits. For example, Xu et al. (2020) demonstrated the potential of integrating speech, video, and text data in behavior prediction [12]. The study by Baltrusaitis et al. (2018) also showed that multimodal data fusion can significantly improve prediction accuracy [13].

Though there is extensive research on predicting student dropout [14], most models rely on online data. Online sys- tems combine the semantic search for analysis and pattern identification for student data, however, semantic search has its limitations [15], [16]. This reliance is understandable, as online data is more structured, larger in volume, and richer in variety, all of which support effective modeling. In contrast, offline educational environments often lack sufficient data in both quantity and quality to train advanced models [17], [18], [19], [20]. As a result, the predictive models currently used in schools are typically basic and often limited to simple regression or statistical summaries, which are disconnected from modern modeling techniques.

In contrast to these simplistic models, educational theory offers a rich, well-developed, and logically coherent foundation for understanding student behavior and outcomes.

However, much of this theoretical insight remains difficult to quantify [21], leaving most predictive models reliant on surface-level observable patterns and lacking a deeper grasp of causal mechanisms. By integrating key elements of educational theory into data modeling, we aim to bridge this gap, extracting deeper, more meaningful features. This approach has the potential to address the limitations of offline data, enabling the development of models that are not only more robust but also offer greater explanatory power for advancing research and practice in education.

Through data analysis and insights derived from the real-world experiences of secondary school teachers, we have discovered that student dropouts are not entirely unpredictable. Sudden changes in student behavior often serve as early indicators of their eventual decision to drop out. Additionally, we have identified several professional papers that explore this phenomenon. Finn and Rock [22] demonstrated that significant shifts in student behavior, such as disengagement or withdrawal, strongly correlate with dropout risk, emphasizing the importance of behavioral stability for academic success. Similarly, Rumberger and Larson [23] highlighted the role of behavioral disruptions—such as irregular attendance and disengagement—as critical precursors to dropout decisions, underscoring the need to monitor these patterns. Jimerson et al. [24] further identified the developmental trajectory of behavioral changes, noting that sudden declines in academic performance or classroom participation often signal underlying issues that lead to dropout. Lee and Burkam [25] confirmed this by showing that behavioral disengagement, including absenteeism and lack of participation, is a strong predictor of dropout in high school settings.

More recent educational studies have provided additional evidence for the relationship between behavioral changes and dropout. Archambault et al. [26] found that early disengagement patterns, particularly in middle and early high school years, are linked to long-term academic failure and dropout. Lessard et al. [27] demonstrated that students who exhibit abrupt behavioral shifts, such as increased defiance or decreased effort, are more likely to leave school prematurely. These findings collectively establish that behavioral volatility is not just a symptom but a precursor of dropout risks, making it a crucial factor for intervention strategies.

Despite extensive theoretical support emphasizing the strong connection between behavioral changes and student dropout, few methods effectively quantify these behavioral transformations [28]. To address this gap, we propose the Dual-Modal Multiscale Sliding Window Model (DMSW), designed for use with offline educational data in real-world settings. This model offers several key contributions:

1) **Introduction of the DMSW for Predicting Student Dropout Risk:** Our model effectively captures student behavior changes by integrating features from both academic performance and behavioral data. By employing multiscale sliding windows, it extracts more discriminative features, achieving a 15% improvement in dropout prediction accuracy compared to baseline models.
2) **Empirical Validation of Behavioral Changes:** We conducted a comprehensive analysis of the model to verify the significant impact of sudden behavioral shifts on dropout risk.
3) **Evaluation of Model Effectiveness:** Through ablation studies and comparative experiments, we demonstrate the necessity and effectiveness of each component within the DMSW framework.

In addition to crafting an efficient model for detecting sudden behavior changes, we investigated several research questions (RQs) to validate our theoretical approach to dropout prediction:

- **RQ1:** Is there a correlation between the behavioral changes identified by the DMSW Model and the occurrence of student dropout?
- **RQ2:** What types of student dropouts can the DMSW Model accurately predict, and in what scenarios does it perform poorly?
- **RQ3:** Based on our observations, what specific behaviors are most likely to lead to dropout?

Our extensive analysis of the model's limitations ultimately identifies behavioral patterns that are strongly associated with a high risk of dropout.

## II. RELATED WORK

Predicting student dropout behavior is an important topic in educational science research. With the advancement of technology, research methods have evolved from traditional statistical analysis to modern deep learning techniques.

*A. Deep Learning Methods*

The prediction of student dropout behavior has been significantly advanced with the emergence of deep learning tech- nologies. Unlike traditional approaches, these methods offer enhanced data processing capabilities and leverage complex model architectures to improve both prediction accuracy and model stability. Among these methods, CNNs stand out due to their success in image processing and feature extraction. For instance, Deng et al. (2017) proposed a CNN-based approach that enhanced prediction performance by automating feature extraction [7]. While CNNs are adept at handling large-scale datasets and capturing complex patterns, their application to student dropout prediction presents certain limitations. Specifically, CNNs are more suited to static image analysis, mak- ing them less effective in modeling dynamic, time-sensitive changes in student behavior, which often require sequential data processing capabilities.

To address the limitations inherent in CNNs, RNNs have been employed due to their ability to process sequential data and model the temporal evolution of student behaviors. Goodfellow et al. (2016) demonstrated the effectiveness of RNNs in modeling time-series data for student prediction tasks, highlighting their strength in capturing behavior changes over time [9]. Moreover, Long Short-Term Memory networks (LSTMs) [29], a variant of RNNs, offer a solution to the long-term dependency issues typical of sequential data [29], [30]. However, despite their potential, LSTMs can face computational challenges, particularly when dealing with lengthy sequences, necessitating further model optimization to maintain efficiency in real-world educational contexts.

In recent years, the application of Graph Neural Networks (GNNs) has introduced a novel perspective to modeling student behavior, especially when analyzing complex relationships in social networks [31]. For example, Zhang et al. (2022) utilized GNNs to investigate the influence of students' social interactions on dropout behavior, achieving noteworthy results [32]. GNNs are well-suited to capture the intricate connections within social data, yet their performance can be hampered by low processing efficiency for large-scale graph structures and the necessity for high-quality graph data, which remains a significant barrier to scalability.

Expanding beyond individual model approaches, multi-modal deep learning methods have emerged as a promising avenue to enhance prediction accuracy by integrating diverse data sources. Xu et al. (2020) illustrated the potential of such approaches by combining voice, video, and text data for student behavior prediction [12]. This comprehensive data integration offers a holistic view of students but brings inherent challenges, including complex data preprocessing and feature alignment. Building on this, Wang et al. (2023) proposed integrating biometric and academic data to further boost model performance [33]. Despite their potential, multimodal approaches must contend with the practical realities of offline data, which primarily consist of textual and numerical records, making synchronization across diverse data modalities challenging.

Given the shortcomings observed in existing methods for capturing behavior changes, we propose a new bimodal model that leverages multi-scale sliding windows and language models. This approach offers a more robust mechanism to improve the prediction of student dropout risk by dynamically adapting to variations in student behavior while addressing the challenges posed by existing methods.

B. Multi-scale Sliding Window

Multi-scale sliding window technology captures changes in student behavior through sliding windows of different time scales. This method can flexibly adapt to different time features, enhancing prediction accuracy. Liu et al. (2021) validated the effectiveness of multi-scale sliding windows in behavior prediction in their study [34]. Multi-scale sliding window technology was initially used in signal processing and has been widely applied in various time-series analysis tasks. For example, Batal et al. (2013) used multi-scale sliding windows in medical data analysis to successfully capture key features of patients' condition changes [35]. Similarly, Cui et al. (2016) applied this technology in financial data analysis, effectively predicting short-term fluctuations in the stock market [36].

In our problem, multi-scale sliding window technology has notably advantages. First, it can capture both short-term and long-term changes in student behavior, which is crucial for predicting sudden behavior changes. Second, this method can handle data of different time scales, allowing the model to adapt to the diversity of student behavior. Finally, multi-scale sliding window technology improves the robustness and prediction accuracy of the model when dealing with complex time-series data.

C. Language Models in Educational Prediction

Language models have proven invaluable for educational prediction tasks, particularly due to their capacity to process and derive meaningful insights from vast volumes of unstructured data. Devlin et al. (2019) introduced the Bidirectional Encoder Representations from Transformers (BERT) model, which has consistently demonstrated exceptional performance across various natural language processing (NLP) tasks by capturing deep contextual relationships within textual data [37]. In educational applications, language models excel at analyzing diverse data sources such as student assignments, communication records, and online discussion forums, thereby providing critical insights into student engagement, compre- hension levels, and potential dropout risks [38], [39].

Liu et al. (2020) utilized transformer-based models to evaluate student performance and engagement in online courses through detailed analysis of discussion forum interactions [40].

Additionally, language models have enhanced the identification of at-risk students by processing qualitative inputs such as teacher comments, feedback, and student essays. Through pre-trained transformers, researchers can capture subtle linguistic signals that may reveal shifts in student behavior, motivation, or understanding [42]. This capacity to detect nuanced variations offers a powerful tool for early intervention strategies, ultimately supporting student success.

Despite their strengths, language models face significant limitations in offline educational settings. Adapting these models to the unique vocabulary and context of educational data is challenging, while their high computational costs and reliance on annotated training data limit accessibility for resource-constrained schools. Instead of relying solely on emotional indicators, focusing on behavior patterns to infer emotional states offers a more practical and robust approach. Developing lightweight, context-aware models capable of analyzing behavioral shifts provides a scalable solution for real-world educational environments [43].

### III. PROPOSED DUAL-MODAL MULTISCALE SLIDING WINDOW (DMSW) MODEL

#### A. Data Source and Data Processing

Our data is the offline educational data that sourced from the original records of diverse schools in Hong Kong. It can be divided into two major categories based on data type: numerical data and textual data. The numerical data primarily records each student's academic performance in three subjects: Chinese, Mathematics, and English. Each subject had six exams throughout the year. Since the maximum score for each exam may vary, we converted each exam score to a relative score based on a 100-point scale and recorded the grade ranking for each exam to facilitate standardized comparisons. The textual data covers detailed descriptions of student behavior at school, including absenteeism, participation in activities, punishments, and rewards. We divided the year into six periods based on exam times and summarized all relevant behaviors in each period into an integrated description. For example, "During this period, the student was absent X times, the reasons including... received Y rewards for reasons such as... faced Z punishments for reasons such as... participated in N activities, including...". For students defined as at-risk, we primarily identified them based on their dropout behavior. During data collection and processing, we strictly adhered to the school's privacy policies and ethical standards to protect students' personal information from being disclosed.

Given that at-risk students constitute a minority of the total student population, our data suffers from significant class imbalance. To address this issue, we employed the Synthetic Minority Over-sampling Technique (SMOTE) [44]. SMOTE is a popular method used to balance datasets by generating synthetic instances of the minority class. By creating new at-risk student records that are similar to existing ones, SMOTE helps to mitigate the imbalance, ensuring that the model can learn effectively from both the majority and minority classes. This balanced data was then used for subsequent analyses.

#### B. Model Description

In this study, we propose a Dual-Modal Multiscale Sliding Window (DMSW) model to predict student dropout risk using the aforementioned data sources. This model is based on students' behavioral patterns and academic performance over a year. By using sliding windows of different sizes, we capture the magnitude and speed of changes in student behavior as new features to predict whether students will choose to drop out. Our model description is divided into five feature extraction, modality fusion, student behavior change feature extraction, and loss function design. The complete model structure is shown in Fig. 1.

*1) Text Feature Extraction:* The original data records student behaviors across four tables, including participation, rewards, punishments, and absenteeism, with detailed timing and descriptions. To extract semantic information, we preprocess the data by aggregating each student's behaviors into descriptive text paragraphs for six time periods over a year. Using the BERT model, we extract high-dimensional semantic features from these texts.

BERT is particularly suited for this task as it captures the nuanced, context-dependent nature of student behavior records, which often include complex descriptions of events. Its bidirectional understanding ensures subtle differences in behaviors are effectively captured, improving differentiation between normal and at-risk students. After extracting embeddings with BERT, we refine them using a Multi-Layer Perceptron (MLP) to better suit dropout prediction. BERT's pretraining on a large corpus also enhances its ability to handle varied language styles in educational records, improving the accuracy of behavior analysis.

*2) Numerical Feature Extraction:* Numerical data, such as students' grades and rankings, is processed using an autoencoder to extract key features. An autoencoder is an unsupervised model that compresses data into a lower-dimensional space and reconstructs it. It consists of an encoder, which maps the input data, and a decoder, which reconstructs it from the compressed space. Autoencoders are particularly effective at feature extraction, compression, and denoising, improving the model's robustness. After obtaining the numerical embeddings, we fine-tune them with an MLP to optimize their use for dropout prediction. Previous studies have demonstrated that autoencoders perform well in educational data processing and prediction tasks.

*3) Modality Fusion:* In the modality fusion part, we directly concatenate the text features and numerical features. This ap- proach allows us to retain all feature information from different modalities, maximizing the complementary characteristics of text and numerical data. Additionally, directly connecting feature vectors is a simple and effective fusion method. It avoids the computational overhead and implementation complexity associated with more intricate fusion strategies. Recent studies also demonstrate that directly connecting multimodal features performs well in various tasks. For instance, some studies in sentiment analysis and multimodal emotion recognition have achieved notable results using direct concatenation methods.

*4) Behavior Change Feature Extraction:* To capture abrupt changes in student behavior and improve dropout prediction accuracy, our study incorporates educational theories that identify behavioral transitions as early indicators of disengagement. Leveraging a multiscale sliding window analysis, the model dynamically tracks behavioral fluctuations over time, capturing both short-term anomalies and long-term trends linked to at-risk behaviors, such as increased absenteeism or sudden performance declines. Grounded in research by Finn and Rock (1997) and Rumberger and Larson (1998), this approach models critical behavioral shifts as precursors to dropout. By extracting multiscale features, the module enhances the model's ability to identify key inflection points, bridging educational theory and data-driven analysis to support early intervention strategies in education.

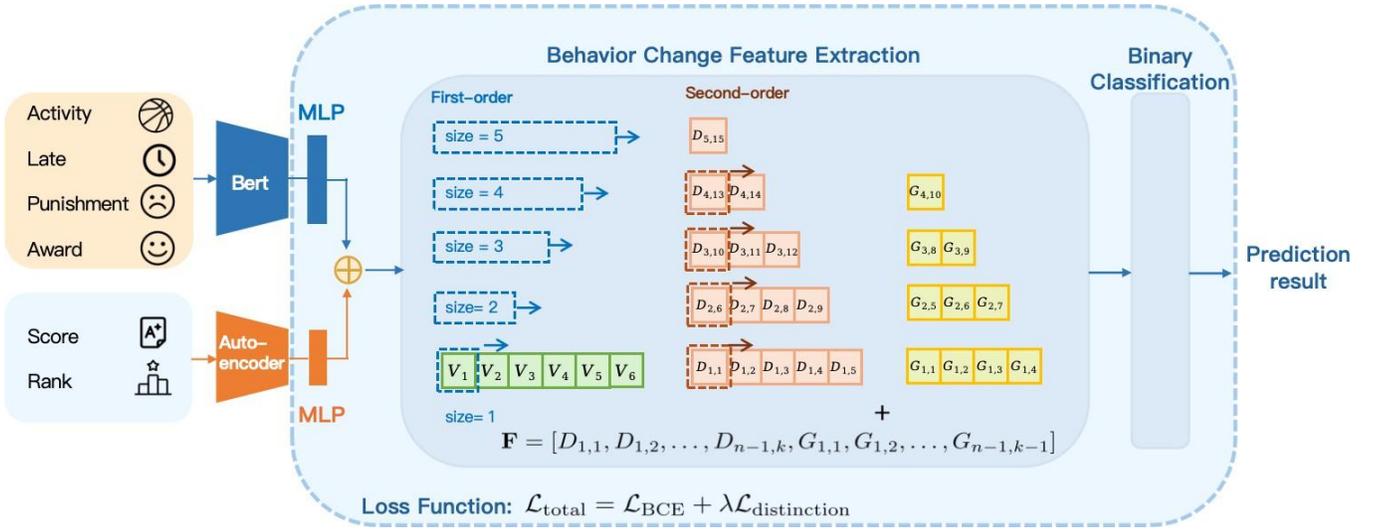

Fig. 1: Dual-Modal Multiscale Sliding Window (DMSW) Model structure

**Sliding Window Method Overview**

The sliding window method segments time-series student behavior data into smaller, overlapping windows to capture changes at different time scales. Let $t_1, t_2, \ldots, t_n$ represent the sequence of time points in the dataset, and let $V_{t_1}, V_{t_2}, \ldots, V_{t_n}$ denote the behavior feature vectors at these respective times. These vectors encode the behavioral characteristics of a student at each time point. The goal of the sliding window method is to analyze this sequence to identify patterns and abrupt changes that may indicate a risk of dropout.

**First-Order Sliding Window Analysis**

The first-order sliding window operates on the original sequence of behavior feature vectors $V_{t_1}, V_{t_2}, \ldots, V_{t_n}$. For a window size $a$ (ranging from 1 to $n-1$), the sliding window moves across the sequence by one unit at a time, generating $n-a$ features for each window size. The features are denoted as $D_{a,1}, D_{a,2}, \ldots, D_{a,k}$, where the first subscript $a$ indicates the window size, and the second subscript represents the feature index.

For a given window size a, the similarity measure between two feature vectors Vti and Vti+a within the window is computed using cosine similarity, as shown in Equation (1):

$$D_{a,i} = \cos(\theta) = \frac{V_{ti} \cdot V_{ti+a}}{\|V_{ti}\| \|V_{ti+a}\|}, \quad i = 1, 2, \ldots, k \quad (1)$$

This process is repeated for all window sizes $a = 1, 2, \ldots, n-1$, resulting in a comprehensive set of first-order features. These features $D_{1,1}, \ldots, D_{n-1,k}$ encode behavioral changes at varying time scales.

**Second-Order Sliding Window Analysis**

The sliding window for second-order analysis operates on the first-order features $D_{a,i}$, focusing strictly on features computed within the same window size $a$. For a fixed window size $a$, the second-order sliding window has a size of 1, meaning it slides over the sequence of first-order features $D_{a,1}, D_{a,2}, \ldots, D_{a,n-a}$. The analysis captures the rate of change or similarity between consecutive first-order features.

For example, given two consecutive first-order features $D_{a,i}$ and $D_{a,i+1}$, the second-order similarity measure $G_{a,i}$ is calculated as follows:

$$G_{a,i} = \cos(\theta) = \frac{D_{a,i} \cdot D_{a,i+1}}{\|D_{a,i}\| \|D_{a,i+1}\|}, \quad i = 1, 2, \ldots, q \quad (2)$$

It is important to note that second-order sliding windows operate strictly within the same window size $a$. For instance, features $[D_{a,1}, D_{a,2}]$ are valid, while comparisons across different window sizes (e.g., $D_{a,1}$ and $D_{b,1}$ for $a \neq b$) are not performed, as such differences lack meaningful interpretation.

After computing all first-order and second-order features, they are concatenated to form the final feature vector **F**, as shown in Equation (3):

$$\mathbf{F} = [D_{1,1}, D_{1,2}, \ldots, D_{n-1,n-1}, G_{1,1}, G_{1,2}, \ldots, G_{n-2,n-2}] \quad (3)$$

This rich feature vector **F** encapsulates both the magnitude and rate of behavior changes across different time scales, providing a robust representation of student behavior dynamics. Finally, we use a two-layer Multi-Layer Perceptron (MLP) as a classifier to process **F**, yielding binary dropout predictions that identify at-risk students with high accuracy.

*5) Loss Function Design:* The loss function for our DMSW model is designed to achieve two primary objectives: accurately

predicting student dropout risk and effectively capturing the impact of abrupt behavioral changes. Given these goals, the loss function comprises two key components: the classification loss and the behavior change distinction loss.

**1. Classification Loss:**

Our model's primary task is to classify whether a student is at risk of dropping out. For this binary classification task, we use Binary Cross-Entropy (BCE) loss, which is well-suited for such tasks. The BCE loss is defined as follows (4):

$$L_{BCE} = -\frac{1}{N}\sum_{i=1}^{N}[y_i\log(p_i) + (1-y_i)\log(1-p_i)] \quad (4)$$

Where
- $y_i$ is the actual label for student $i$ (1 if the student is at risk, 0 otherwise),
- $p_i$ is the predicted probability that student $i$ is at risk,
- $N$ is the total number of students.

This loss function penalizes incorrect predictions, ensuring that the model precisely classifies students based on their risk of dropping out.

**2. Behavior Change Distinction Loss:**

To ensure that the model pays special attention to students with significant behavioral changes, we introduce a behavior change distinction loss. This loss amplifies the model's sensitivity to students whose behavior exhibits abrupt and significant changes, as these students are more likely to be at risk of dropping out.

Given that our model uses multiscale sliding windows to capture first-order and second-order differences in student behavior over time, we define the behavior change distinction loss as follows (5), (6):

$$L_{distinction} = \frac{1}{M}\sum_{j=1}^{M}\sum_{i=1}^{N}[y_i \cdot \Delta_j - (1-y_i) \cdot \Delta_j] \quad (5)$$

$$\text{where } \Delta_j = \max(0, D_j - \delta). \quad (6)$$

where:
- $D_j$ represents the magnitude of behavior change (e.g., first-order or second-order difference) for the jth feature,
- $\delta$ is a threshold value used to distinguish significant behavior changes. In our case, $\delta$ is dynamically set as the 85th percentile (top 15%) of all behavior changes within the current window size and position, ensuring that the model focuses on the most significant behavior changes,
- $M$ is the number of behavior change features extracted through the sliding window analysis.

The behavior change distinction loss, $L_{distinction}$, provides key benefits in predicting student dropout risk. By focusing on significant behavior changes — those above the 85th percentile within the current window — the loss function ensures the model is sensitive to dramatic shifts in student behavior. This helps identify at-risk students, who typically show noticeable changes before dropping out. The dynamic threshold $\delta$, based on the top 15% of behavior changes, allows the model to adapt to varying levels of change across students and time. Moreover, by distinguishing between major and minor changes, $L_{distinction}$ reduces overfitting, preventing the model from overemphasizing insignificant fluctuations. This targeted approach enhances the model's generalizability and accuracy in capturing dropout indicators.

**3. Total Loss Function:**

The overall loss function is a weighted combination of the classification loss and the behavior change distinction loss shown as (8):

$$L_{total} = L_{BCE} + \lambda L_{distinction} \quad (7)$$

where $\lambda$ is a hyperparameter that controls the balance between predicting dropout risk and emphasizing behavior changes. In our experiments, we set $\lambda = 0.5$, which we manually tuned as the optimal parameter to ensure a balance between the Classification Loss and the Behavior Change Distinction Loss. This setting ensures that the model effectively captures the importance of both components, providing an indepth approach to predicting student dropout.

## I. EXPERIMENT RESULTS AND DISCUSSIONS

In this section, we analyze and evaluate the potential of our proposed model using real data. Our experiment consists of two parts. In the first part, Learning Analysis, we address the research questions posed earlier to demonstrate the feasibility of our DMSW model, analyze its limitations, and provide data-driven insights that teachers can use as a reference. In the second part, we conduct an efficacy analysis of the model. By comparing it with baseline models, we find that our proposed model outperforms existing student dropout prediction approaches. Additionally, through ablation studies, we confirm the necessity and contribution of each component of our model.

### A. Dataset

Our dataset is sourced from the original records of a secondary school in urban Hong Kong, covering detailed records of 1,721 students over one year between 2019 and 2022. We assume that student behavioral patterns do not dramatically vary across different years. Based on our definition of at-risk students, we classified 210 students as at-risk, specifically those who dropped out within the following year.

### B. Learning analysis

*1) RQ1: Is there a correlation between the behavioral changes identified by the DMSW Model and the occurrence of student dropout?*

To explore the relationship between student behavioral changes and dropout in depth, we presented the impact of behavioral changes on dropout prediction through specific statistical results and case studies. The Ordinary Least Squares (OLS) regression analysis was used to model the relationship between behavioral changes and dropout risk. OLS is a statistical method for estimating the parameters of a linear regression model, aiming to minimize the sum of the squared differences between observed and predicted values. It helps in understanding how changes in predictor variables are associated with changes in the outcome variable.

From the OLS regression results, shown in Table I, the model's R-squared value is 0.757, indicating that it explains about 75.7% of the variance in dropout risk. This high R-squared value demonstrates the model's good fit. Additionally, the F-statistic is 63.10 with a p-value of 2.49e-151, indicating that the model is highly relevant overall, meaning that these predictor variables together substantially explain the variance in dropout risk.

After grouping the variables by different window sizes, the table further presents the average coefficients (Coef), standard errors (Std Err), and p-values for each group. Overall, the first-order groups with smaller window sizes (e.g., window sizes 1 and 2) generally exhibit significant positive effects, indicating that an increase in these variables substantially raises the likelihood of an individual being labeled as at risk. These groups tend to have relatively large coefficients and low p-values, highlighting their importance in the prediction model. Although the second-order groups also demonstrate significant effects in some cases, they generally have lower coefficients and are less significant than the first-order groups. Among the groups with larger window sizes (e.g., window sizes 3, 4, and 5), some groups also display significant positive effects. In particular, the first-order groups for window sizes 3 and 5 have larger coefficients and stronger significance levels. Overall, groups with higher significance contribute more to the predictive performance of the model. This is especially evident for the lower-order groups with larger window sizes, where changes in these variables have a more pronounced impact on the at-risk label.

*2) RQ2: What types of student dropouts can the DMSW Model accurately predict, and in what scenarios does it perform poorly?*

In this section, we analyze the prediction accuracy and examine instances of incorrect predictions. By summarizing the characteristics of students with accurate predictions, we highlight the key information identified by the model, offering practical insights for teachers' observations. For incorrect predictions, we evaluate the model's weaknesses, identifying areas for improvement and providing a foundation for future research.

To quantify student performance, we use the composite metric combining Z-scores and relative rankings:

$$\text{Composite Value} = \omega_1 \times Z + \omega_2 \times \left(1 - \frac{\text{Rank} - 1}{\text{Total Students} - 1}\right) \quad (8)$$

Here,

$$Z = \frac{\text{Score} - \text{Mean}}{\text{Standard Deviation}} \quad (9)$$

measures standardized performance, while the ranking term normalizes positional information on a scale of 0 to 1. $\omega_1$ and $\omega_2$ are weights for Z-scores and rankings, respectively, reflecting their relative importance.

TABLE I: OLS Regression Results (Averaged by Window Size Groups)

| Dep. Variable: | AT-RISK LABEL | | |
|---|---|---|---|
| Statistic | Value | Statistic | Value |
| R-squared | 0.757 | Adj. R-squared | 0.751 |
| F-statistic | 63.10 | Prob (F-statistic) | 2.49e-151 |
| **Coefficients (Selected):** | | | |
| Window Size Group | Coef | Std Err | P>\|t\| |
| Window Size 1, 1st order (x1 - x5) | 6.0155 | 3.4282 | 0.0281 |
| Window Size 1, 2nd order (x6 - x9) | 3.7151 | 1.2176 | 0.0593 |
| Window Size 2, 1st order (x10 - x13) | 5.2346 | 2.5705 | 0.0200 |
| Window Size 2, 2nd order (x14 - x16) | 1.4057 | 0.8818 | 0.0390 |
| Window Size 3, 1st order (x20 - x21) | 8.4057 | 2.8818 | 0.0390 |
| Window Size 3, 2nd order (x17 - x19) | 1.3833 | 1.2182 | 0.4347 |
| Window Size 4, 1st order (x22 - x23) | 3.1923 | 1.1290 | 0.0341 |
| Window Size 4, 2nd order (x24) | 2.0866 | 2.1240 | 0.0609 |
| Window Size 5, 1st order (x25) | 7.5674 | 2.8310 | 0.0240 |

*a) POSITIVE EXAMPLES:* Based on the data provided for Student A in Table II and Fig. 2a, we can identify the main factors contributing to their dropout risk. Firstly, the student showed a significant increase in absenteeism during the fifth period, which was followed by a noticeable decline in academic performance, particularly in Chinese and Mathematics. Although the student received a positive behavior record in the sixth period, it was not enough to offset the severe negative impact of frequent absences and declining grades. Overall, frequent absenteeism and academic decline were the primary factors leading to the student's eventual dropout.

Based on the data provided for Student B in Table III and Fig. 2b, the student performed well in earlier semesters and even received rewards for good behavior during the fourth and fifth periods. However, in the sixth time period, the student was severely reprimanded for repeatedly inappropriately touching a female group mentor. This serious behavioral issue not only led to potential further disciplinary actions by the school but also had a negative impact on the student's psychological state and academic performance. As seen in the grade char the student's scores in Chinese, Mathematics, and English were all significantly affected. Therefore, we can conclude that the primary cause of Student B's dropout was the severe behavioral issue in the sixth period and the subsequent chain reaction it triggered.

In the positive case section, we extracted two representative examples from a large number of cases and identified two main reasons for student dropout: On one hand, students who drop out in the second year typically experience significant absenteeism in the previous year, accompanied by a noticeable decline in academic performance. On the other hand, when students face severe reprimands, it can lead to a significant drop in their overall grades. In the following chapters, we will calculate the

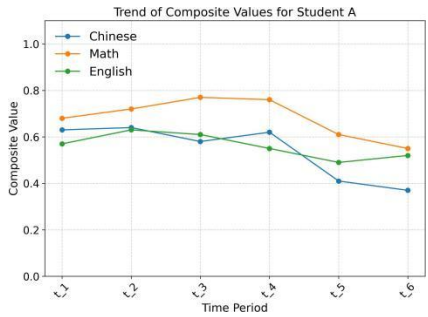
(a) The grade records of Student A.

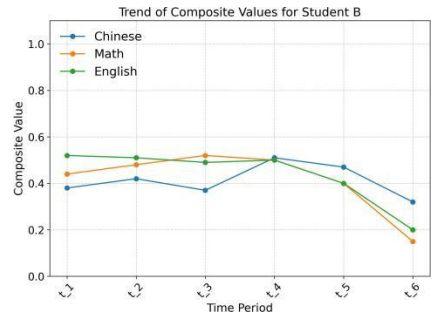
(b) The grade records of Student B.

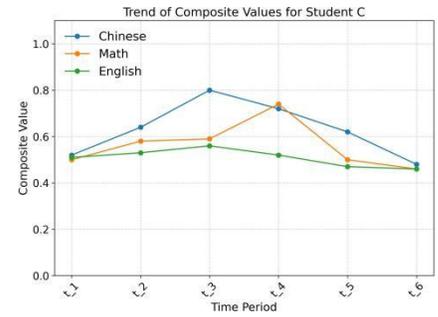
(c) The grade records of Student C.

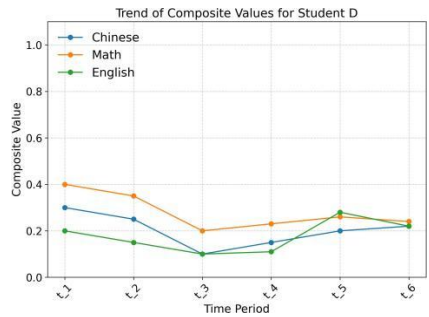
(d) The grade records of Student D.

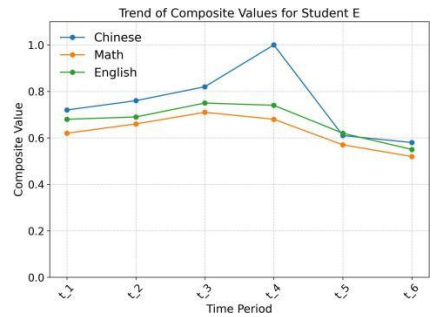
(e) The grade records of Student E.

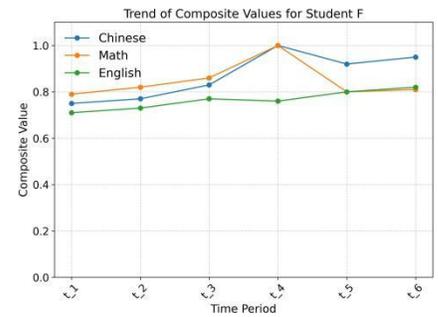
(f) The grade records of Student F.

Fig. 2: The grade records of students.

TABLE II: Positive example 1: Behavior Records for Student A

| Time Period | Activity Records | Punishment Records | Reward Records | Absence Records |
|---|---|---|---|---|
| 1 | / | / | / | / |
| 2 | / | / | / | / |
| 3 | / | / | / | 1 absence record: All day (sick leave) |
| 4 | / | / | / | / |
| 5 | / | / | / | 24 absence records: PM, PM, AM, PM, AM, PM, AM, PM, AM, PM, AM, PM, AM, PM, AM, PM, AM, PM, AM, PM, AM, AM, PM, AM (sick leave, parental leave, other reasons, late) |
| 6 | / | / | 1 reward record: good behavior 1 time(s). Reason: None | 2 absence records: AM, All day (other reasons, late, transportation issues) |

TABLE III: Positive example 2: Behavior Records for Student B

| Time Period | Activity Records | Punishment Records | Reward Records | Absence Records |
|---|---|---|---|---|
| 1 | / | / | / | / |
| 2 | / | / | / | / |
| 3 | / | / | / | / |
| 4 | / | / | 1 reward record: good behavior 1 time(s). Reason: None | / |
| 5 | / | / | 1 reward record: good behavior 1 time(s). Reason: None | / |
| 6 | / | 1 punishment record: deliberate touching of female group mentor. Type: severe reprimand 1 time(s) | 2 reward records: good behavior 1 time(s). Reason: None, good behavior 1 time(s). Reason: None | / |

TABLE IV: Counter-example 1: Behavior Records for Student C

| Time Period | Activity Records | Punishment Records | Reward Records | Absence Records |
|---|---|---|---|---|
| 1 | / | / | / | 1 absence record: AM (late, reason: overslept) |
| 2 | / | / | / | / |
| 3 | / | / | / | 2 absence records: All day, All day (late, reasons: overslept, other reasons (exempted)) |
| 4 | / | / | 4 reward records: merit 1 time(s), good behavior 1 time(s), good behavior 1 time(s), good behavior 1 time(s). Reasons: None, None, None, None | / |
| 5 | / | / | 5 reward records: good behavior 1 time(s), good behavior 1 time(s), merit 1 time(s), good behavior 1 time(s), merit 1 time(s). Reasons: None, None, None, None, None | 1 absence record: AM (late, reason: overslept) |
| 6 | / | 1 punishment record: verbally abusing teacher and throwing objects in class, showed remorse after. Type: minor offense 1 time(s) | / | 6 absence records: AM, PM, AM, PM, AM, PM (reasons: parental leave, parental leave, parental leave, parental leave, parental leave, parental leave) |

TABLE V: Counter-example 2: Behavior Records for Student D

| Time Period | Activity Records | Punishment Records | Reward Records | Absence Records |
|---|---|---|---|---|
| 1 | / | / | / | / |
| 2 | / | / | 1 reward record: merit 1 time(s). Reason: None | / |
| 3 | / | / | / | 16 absence records: All day, All day, All day, All day, All day, All day, All day, All day, All day, All day, All day, All day, All day, All day, All day, All day (reasons: unknown, unknown, unknown, unknown, unknown, unknown, unknown, unknown, unknown, unknown, unknown, unknown, unknown, unknown, unknown, other reasons) |
| 4 | / | / | / | 12 absence records: All day, All day, All day, All day, All day, All day, All day, All day, All day, All day, All day, All day (reasons: unknown, unknown, unknown, unknown, unknown, unknown, unknown, unknown, unknown, unknown, unknown, unknown) |
| 5 | / | / | 4 reward records: good behavior 1 time(s), good behavior 1 time(s), good behavior 1 time(s), good behavior 1 time(s). Reasons: None, None, None, None | 2 absence records: AM, PM (reasons: other reasons, other reasons) |
| 6 | / | / | 1 reward record: good behavior 1 time(s). Reason: None | 6 absence records: AM, PM, AM, PM, AM, PM (reasons: parent did not answer phone, parent did not answer phone, other reasons, other reasons, parental leave, parental leave) |

TABLE VI: Counter-example 3: Behavior Records for Student E

| Time Period | Activity Records | Punishment Records | Reward Records | Absence Records |
|---|---|---|---|---|
| 1 | / | / | 1 reward record: good behavior 1 time(s). Reason: None | / |
| 2 | / | / | / | / |
| 3 | / | / | 3 reward records: good behavior 1 time(s), merit 1 time(s), good behavior 1 time(s). Reasons: None, None, None | / |
| 4 | / | / | 2 reward records: good behavior 1 time(s), good behavior 1 time(s). Reasons: None, None | / |
| 5 | / | / | 3 reward records: good behavior 1 time(s), good behavior 1 time(s), good behavior 1 time(s). Reasons: None, None, None | 2 absence records: AM, PM (reasons: sick leave, sick leave) |
| 6 | / | / | 2 reward records: good behavior 1 time(s), good behavior 1 time(s). Reasons: None, None | / |

TABLE VII: Counter-example 4: Behavior Records for Student F

| Time Period | Activity Records | Punishment Records | Reward Records | Absence Records |
|---|---|---|---|---|
| 1 | / | / | / | 1 absence record: AM (sick leave) |
| 2 | / | / | / | / |
| 3 | / | / | 1 reward record: good behavior 1 time(s). Reason: None | 1 absence record: All day (sick leave) |
| 4 | / | / | / | / |
| 5 | / | / | / | 2 absence records: AM, PM (reasons: sick leave, sick leave) |
| 6 | / | / | 1 reward record: good behavior 1 time(s). Reason: None | / |

probability of these situations occurring to determine whether they are common trends. The next section will discuss cases of incorrect predictions made by the model.

*b) COUNTER EXAMPLES:* This section is divided into two categories: one involves incorrect predictions of high-risk students (predicted as high-risk but actually non-high-risk), and the other involves incorrect predictions of non-high-risk students (predicted as non-high-risk but actually high-risk). Each category will present two examples, and the section will conclude with a summary of the reasons for the errors.

Based on the data provided for Student C in Table IV and Fig. 2c, the student had multiple instances of absenteeism during the sixth period, with the reason being parental leave. Additionally, the student was praised multiple times for good behavior during the fourth and fifth periods, but in the sixth period, the student was reprimanded for insulting the teacher and throwing objects in class. Overall, the student's performance showed a slight decline. Specifically, the student's grades dropped after the absences, but the limited and insufficientlysignificant features led to an incorrect prediction by the model.

According to the data shown in Table V and Fig. 2d, Student D started accumulating a significant number of absences from the third period, but received some awards in the last two periods. The grades fluctuated at a consistently low level. The combination of frequent absences and low, fluctuating grades might lead the model to incorrectly classify the student as a dropout. However, from a practical perspective, the frequent absences could be due to uncontrollable factors, and while the student's grades are low, they do not show significant fluctuation, indicating that the student may not necessarily be in a passive state.

Here are two examples of incorrectly predicting non-at-risk students. A student (referred to as Student E) analysis of the behavior data Table VI and academic performance Fig. 2e trends reveals that, despite no notable behavioral changes recorded in the data, the student's grades showed a marked decline, especially in the final time period. While the student received numerous commendations in earlier time periods and had no pronounced punishment records, these rewards and absence records did not fully capture the underlying behavioral changes and psychological state. Despite the presence of commendations, the details of behavioral changes may have been inadequately recorded, leading to the model's failure to precisely capture these subtle changes. Consequently, the student's eventual dropout indicates that the model has limitations in capturing sufficiently detailed behavioral changes and maintaining complete records. Based on the provided Table VII and Fig. 2f, this student (referred to as Student F) was predicted by the model not to drop out, but in reality, the student did drop out. The charts show that while the student's grades fluctuated significantly in some time periods, they tended to stabilize overall. However, there is a contradictory pattern between multiple absence records and commendation records in the behavioral data. The student displayed good behavior and received commendations in certain time periods, yet also had multiple absences, primarily due to illness. This conflicting information may have led to the model's failure to clearly capture the student's dropout risk.

As shown in Table VIII, the causes of errors and suggested solutions are summarized. The inaccuracies in the model's predictions may stem from incomplete or insufficiently detailed behavioral data, which hampers the identification of behavioral changes. To enhance the model's predictive capability, a broader range of data (such as psychological data or additional behavioral data) and a longer observation period are needed.

*3) RQ3: Based on our observations, what specific behaviors are most likely to lead to dropout?*

Through the analysis of all available data, we have identified specific behavioral patterns that serve as critical warning signs for student dropout. The following are the key findings:

**Sudden Increase in Absenteeism Accompanied by Significant Academic Decline.** We found that when a student's absenteeism increases by more than five times in one or two time periods, coupled with a decline in academic performance of more than 30%, the likelihood of dropout increases. The overall data indicate that under normal circumstances, the dropout rate is approximately 12.2%. However, when absenteeism spikes and academic performance declines, this dropout rate surges to 68.5%, an increase of 56.3 percentage points.

It is important to note that while this behavioral pattern greatly elevates the risk of dropout, more awards can mitigate this increase. For instance, if a student receives an average of five rewards during this period, the dropout rate decreases from 68.5% to 22.6%. Additionally, we found that the reasons for absenteeism (such as illness or family issues) do not significantly impact the dropout rate, whereas the type of reward does have an obvious effect. Specifically, we categorized rewards into three types: academic competition awards (e.g., winning a contest), academic attitude rewards (e.g., completing assignments on time, no tardiness or absences), and good behavior rewards (e.g., general good conduct without specific reasons). The analysis of the impact

TABLE VIII: Summary of Prediction Errors and Solutions

| Error Situation | Cause | Solution |
|---|---|---|
| Actual 0 predict 1 | The student may have faced external factors leading to unavoidable absences, but their learning attitude remained positive | Collect more extensive features; Include richer psychological characteristics |
| Actual 1 predict 0 | Behavioral data showed no significant changes, and grades declined, but not prominently | Collect more extensive features; Extend the recording period |

*1 indicates dropout, while 0 indicates no dropout.

of different reward types on dropout rates is summarized in Table IX. For instance, the "Academic Competition" reward type, with a dropout rate of 26.5%, reflects the probability of students dropping out when they received this type of reward, particularly in cases where there was a significant increase in absences (more than five additional absences in one or two time periods) and a concurrent academic performance decline of 30%. Similarly, the "Academic Attitude" and "Good Behavior" reward types correspond to dropout probabilities of 20.5% and 28.5%, respectively, under similar conditions. From the comparison, it's evident that rewards focused on academic attitude lead to the lowest dropout rates, while good behavior rewards result in slightly higher dropout rates. This suggests that while all reward types positively influence dropout rates, fostering academic attitudes has the most substantial impact.

**Severe Punishments Leading to Academic Decline and Increased Dropout Rates**. For students who received severe punishments, we observed a 79.7% probability of a subsequent academic decline of over 30%, with 73.9% of these students ultimately choosing to drop out. Through clustering analysis of specific punishment reasons, we categorized them into four types: academic punishments (e.g., repeatedly failing to submit assignments, tardiness, inattentiveness in class), misbehavior punishments (e.g., vandalism, failure to return school property), academic dishonesty punishments (e.g., plagiarism, forging parental signatures), and infringement punishments (e.g., verbal abuse, harassment).

Table IX illustrates the influence of various punishment types on dropout rates. The average dropout rate across all punishment categories is 73.9%. Specifically, "Academic Punishments" and "Misbehavior Punishments" have slightly lower dropout rates of 68.9% and 65.3%, respectively. However, more severe forms of misconduct, such as "Dishonesty" and "Infringement" punishments, are associated with significantly higher dropout rates of 80.6% and 88.4%, respectively. These findings indicate that the type of punishment a student receives can greatly influence their likelihood of dropping out, with more severe infractions, particularly those involving infringement behavior, leading to a substantially higher risk of dropout.

These findings underscore that severe punishments not only lead to significant academic decline but also greatly increase the risk of dropout, particularly in cases involving infringement behavior. This analysis allows for a more insightful identification of behaviors that may lead to student dropout.

### C. Model Structural effectiveness

To demonstrate the effectiveness of our model, we first validated its competitiveness against baseline models. Additionally, to ensure the optimal structure of our model, we focused on two key aspects: comparing the effects of different modalities' fusion positions and varying sliding window sizes, while providing detailed explanations for the results obtained.

TABLE IX: Impact of Different Reward and Punishment Types on Dropout Rates

| Type | Dropout Rate | Change |
|---|---|---|
| **Rewards** | | |
| No Rewards | 68.7% | - |
| Academic Competition | 26.5% | -42.2% |
| Academic Attitude | 20.5% | -48.2% |
| Good Behavior | 28.5% | -40.2% |
| **Punishments** | | |
| Average | 73.9% | - |
| Academic Punishment | 68.9% | -5% |
| Misbehavior | 65.3% | -8.6% |
| Dishonesty | 80.6% | +6.7% |
| Infringement | 88.4% | +14.5% |

This table spresents the impact of various types of rewards and punishments on student dropout rates. Rewards, such as academic competition and good behavior, significantly reduce dropout rates, while certain punishments, like dishonesty and infringement, are associated with increased dropout rates.

*1) Model Performance:* We began by comparing our model with five commonly used models for predicting student dropout in offline educational settings. Specifically, we divided the comparisons into two categories: the first category retained the bimodal aspect of our model, extracting semantic information from text data, but removed the complete sliding window structure, directly feeding the fused features into five different classification models. The second category used a direct numerical approach, where behavioral and textual information was combined with grades and frequencies.

From Table X, bimodal methods significantly outperform single-modality approaches. Combining semantic information from text data with behavioral data consistently improved performance across all baseline models, demonstrating that multimodal data captures richer insights and enhances prediction accuracy.

Our DMSW model excelled across all evaluation metrics, particularly in F1-Score, outperforming other methods by 15%,

TABLE X: Dropout Prediction Performance Comparison

| Model | Accuracy | Precision | Recall | F1-Score |
|---|---|---|---|---|
| Logistic Regression_num | 0.623 | 0.601 | 0.608 | 0.604 |
| SVM_num | 0.645 | 0.622 | 0.630 | 0.626 |
| Decision Tree_num | 0.591 | 0.576 | 0.582 | 0.579 |
| Random Forest_num | 0.675 | 0.653 | 0.660 | 0.656 |
| MLP_num | 0.700 | 0.678 | 0.685 | 0.682 |
| Logistic Regression_bi | 0.782 | 0.721 | 0.730 | 0.725 |
| SVM_bi | 0.793 | 0.734 | 0.742 | 0.738 |
| Decision Tree_bi | 0.816 | 0.729 | 0.715 | 0.712 |
| Random Forest_bi | 0.825 | 0.739 | 0.745 | 0.742 |
| MLP_bi | 0.801 | 0.727 | 0.735 | 0.731 |
| DMSW (Our proposed) | **0.872** | **0.761** | **0.763** | **0.893** |

As shown in the table, models labeled with "num" indicate the outcomes from single modality processing, while those labeled with "bi" indicate the outcomes from bimodal processing. The DMSW model performed the best across all metrics, proving the effectiveness and superiority of our approach.

effectively identifying at-risk students while achieving an optimal balance between precision and recall. Key factors driving this success include: Effective Multimodal Fusion, which integrates semantic and behavioral data to provide a holistic view of student status, overcoming single-modality limitations; and the Sliding Window Technique, which captures temporal variations in behavior critical for early risk detection. Optimal window size selection further boosted performance.

In summary, our study shows that multimodal fusion and structural optimization can significantly improve dropout prediction accuracy. Future work could explore new fusion methods and optimization strategies to further enhance real-world educational applications.

*2) Model Explanation:* In this section, we analyze the impact of the sliding window module's position and size on model performance. First, we explored the position of the sliding window by setting two configurations: applying the sliding window before feature fusion and after feature fusion. As shown in Fig. 3, applying the sliding window after feature fusion more effectively captures the dynamic changes in the time series, thereby enhancing the model's predictive performance. This is because applying the sliding window after feature fusion can integrate the temporal variations of different features, allowing the model to adapt more comprehensively to data changes during feature extraction.

After determining the optimal position of the sliding window, we conducted further research by varying the sliding win- dow size and combination to investigate its impact on model performance. We experimented with various sliding window size configurations, ranging from small to large windows, and conducted a systematic comparative analysis. The results, shown in Fig. 4, indicate that medium-sized sliding windows slightly outperform large windows in balancing feature capture and computational efficiency. Small windows, however, perform the worst as they tend to amplify noise, leading to less reliable predictions. While medium windows capture features effectively with greater precision than large windows, they also maintain model robustness without the drawbacks of noise amplification seen in smaller windows. The medium size provides an optimal balance, offering detailed information without overwhelming the model with irrelevant data, whereas large windows, though stable, may overlook subtle but significant temporal variations.

To comprehensively validate our model design, we also conducted experiments with different combinations of sliding window sizes. We applied different sizes of sliding windows at various feature levels to further enhance the model's performance. Through these combination experiments, we found that multi-level, multi-scale sliding window applications can maintain stable predictive performance across a broader range of data variations.

Specifically, Fig. 5 presents the experimental results of different combinations of sliding window sizes. The last bar in each combination represents the average performance of that combination. Multi-scale sliding window applications have great advantages:

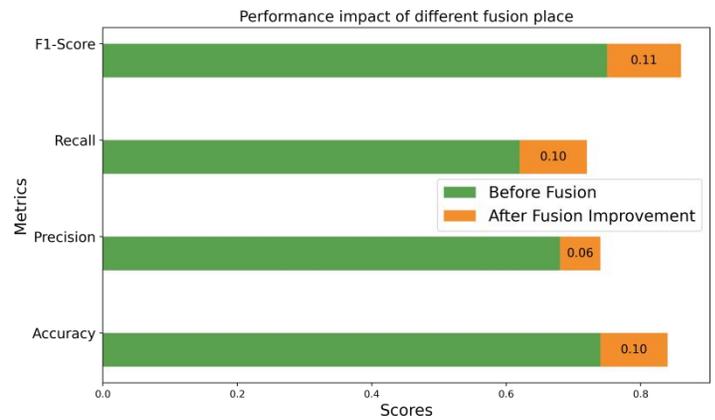

Fig. 3: This horizontal bar chart compares model performance on four metrics (Accuracy, Precision, Recall, F1-Score) before and after applying a sliding window with fusion. The green bars represent the scores before fusion, while the orange segments indicate the improvement gained after fusion. This highlights the positive impact of the fusion process across all evaluated metrics.

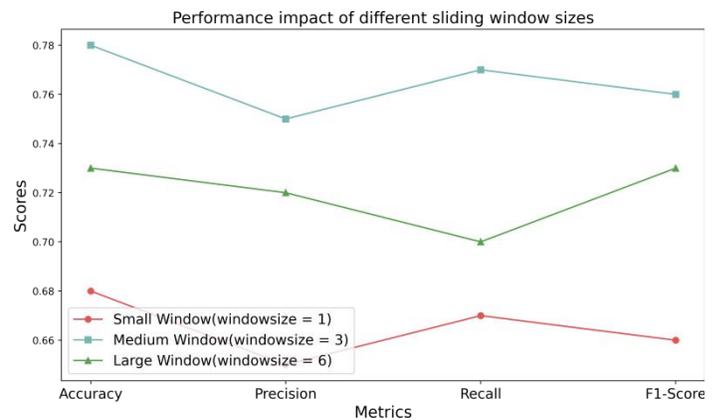

Fig. 4: Large windows are window size 6, medium windows are window size 3, and small windows are window size 1.

- **Capturing Diverse Educational Patterns**: The multiscale sliding window approach enables the identification of both short-term behaviors, such as sudden drops in engagement, and long-term trends, like consistent performance decline. This ability to analyze patterns across different time scales provides educators with deeper insights into students' academic trajectories, allowing for more precise and timely interventions.
- **Supporting Data-Driven Decisions**: By analyzing data at multiple scales, the sliding window method reduces the risk of missing critical behavioral shifts or overemphasizing minor fluctuations. This robustness helps educators make reliable decisions based on stable and meaningful patterns, ensuring that interventions are appropriately targeted.

- **Enhancing Early Prediction for Intervention**: The adaptability of multi-scale sliding windows to complex time-series data ensures high predictive accuracy, even when student behaviors vary widely. This enhances educators' ability to identify at-risk students early, enabling timely support and fostering improved educational outcomes.

Through these combination experiments, we validated the effectiveness of multi-scale sliding window applications, proving that they can enhance the model's predictive performance in complex data environments. In summary, the strategy of combining multi-modal fusion with multi-scale sliding windows provides an effective solution for improving model performance.

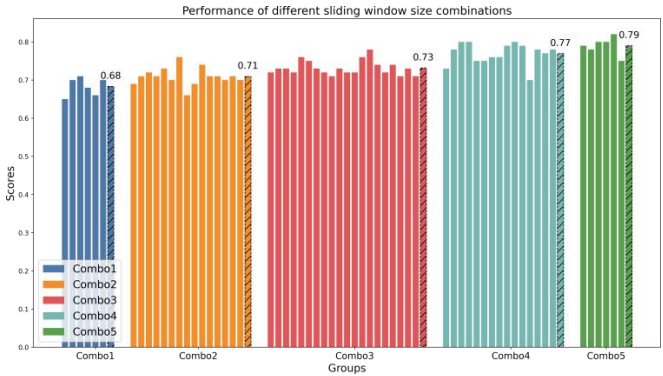

Fig. 5: Combo1 denotes using only one window size, Combo2 denotes using two window sizes, and so on. The bar with stripes represents the average accuracy of the current combo.

This experiment provided a detailed comparative analysis, verifying the great impact of the sliding window module's position and size on model predictive performance. The results show that applying the sliding window after feature fusion and using medium-sized sliding windows perform best in balancing feature capture and computational efficiency. Additionally, the combination of multi-scale sliding windows further enhances the model's robustness and predictive accuracy, offering more reliable technical support for dropout prediction. Future research can continue to explore more modalities of fusion and optimization strategies to achieve better application effects in practical educational environments.

## IV. Deployment

The model discussed in this paper has been successfully integrated into the at-risk student prediction module of a data management system. The system is now in real-world use, providing early prediction capabilities for K-12 teachers, and has received positive feedback. By integrating advanced predictive models, the system helps educators identify students at risk of dropping out in real time, providing data-driven insights for timely intervention and support. For specific positive user experiences from multiple schools, please refer to the conference paper [45].

The deployment of the system enables teachers to identify students' academic issues and behavioral changes earlier, allowing for targeted support measures. This not only enhances educators' ability to track student dynamics but also helps schools improve student retention rates and academic success. With this intelligent prediction tool, teachers can provide more personalized teaching and guidance, significantly improving the equity and effectiveness of educational services.

## V. Conclusion

In this article, a novel Dual-Modal Multiscale Sliding Window (DMSW) model is proposed for predicting student dropout risk by analyzing abrupt behavioral changes and academic performance dynamics. This model addresses the urgent need for predictive tools that operate effectively in offline educational environments, which often lack access to extensive and structured online data. Our five-part architecture includes student behavior text feature extraction, numerical feature extraction of academic performance, modality fusion, behavioral change feature extraction, and a carefully designed loss function. A key highlight of this research is the development of the DMSW model, which integrates semantic text analysis and numerical feature extraction within a dual-modal framework. By employing multiscale sliding windows, the model captures both short-term and long-term behavioral variations, resulting in a 15% improvement in prediction accuracy compared to traditional methods. This improvement is based on a three-year dataset from 1,721 students across three subjects in a secondary school. To our knowledge, this is the first study to systematically incorporate multiscale behavioral dynamics and academic performance features in a real-world educational setting using a dual-modal machine learning approach.

In addition, our research addresses several critical questions: 1) Is there a correlation between the behavioral changes identified by the DMSW Model and the occurrence of student dropout? 2) What types of student dropouts can the DMSW Model accurately predict, and in what scenarios does it perform poorly? 3) Based on our observations, what specific behaviors are most likely to lead to dropout?

The analysis conducted with the DMSW model provides significant insights into predicting student dropout by identifying and quantifying abrupt behavioral changes. The model effectively highlights that these behavioral shifts, particularly when combined with severe punishments, considerably increase the risk of students leaving school. Furthermore, it demonstrates a marked improvement in prediction accuracy over traditional methods, affirming its efficacy in the early identification of at-risk students. These findings underscore the importance of monitoring behavioral changes as a proactive strategy to prevent student dropout. The DMSW model has been successfully integrated into a real-world application system utilized by multiple schools, receiving positive feedback from users. Looking ahead, it can be incorporated into real- time decision support systems based on the current platform, leveraging a data-driven approach to foster supportive learning environments that keep students en-gaged and enable them to reach their full potential within a nurturing educational framework.